\documentclass[sigconf,screen]{acmart}
\copyrightyear{2019} 
\acmYear{2019} 
\setcopyright{rightsretained} 
\acmConference[WSDM '19]{The Twelfth ACM International Conference on Web Search and Data Mining}{February 11--15, 2019}{Melbourne, VIC, Australia}
\acmBooktitle{The Twelfth ACM International Conference on Web Search and Data Mining (WSDM '19), February 11--15, 2019, Melbourne, VIC, Australia}
\acmDOI{10.1145/3289600.3290999}
\acmISBN{978-1-4503-5940-5/19/02}

\usepackage{booktabs} 
\usepackage{xcolor}

\newcommand{\youtube}{YouTube}
\newcommand{\eq}{Equation }
\newcommand{\eg}{\emph{e.g.}, }

\begin{document}
\title{Top-$K$ Off-Policy Correction\\for a REINFORCE Recommender System}

\author{Minmin Chen$^*$, Alex Beutel$^*$, Paul Covington$^*$, Sagar Jain, Francois Belletti, Ed H. Chi}
\thanks{$^*$ Authors contributed equally.}
 \affiliation{%
   \institution{Google, Inc.}
   \city{Mountain View}
   \state{CA}
   \postcode{94043}
 }
\email{minminc,alexbeutel,pcovington,sagarj,belletti,edchi@google.com}

\begin{abstract}
    Industrial recommender systems deal with \emph{extremely large} action spaces -- many millions of items to recommend. Moreover, they need to serve billions of users, who are unique at any point in time, making a complex user state space. 
    Luckily, huge quantities of logged implicit feedback (\eg user clicks, dwell time) are available for learning. Learning from the logged feedback is however subject to biases caused by only observing feedback on recommendations selected by the previous versions of the recommender. In this work, we present a general recipe of addressing such biases in a production top-$K$ recommender system at \youtube, built with a policy-gradient-based algorithm, i.e. REINFORCE \cite{williams1992simple}. The contributions of the paper are: (1) scaling REINFORCE to a production recommender system with an action space on the orders of millions; (2) applying off-policy correction to address data biases in learning from logged feedback collected from multiple behavior policies; (3) proposing a novel top-$K$ off-policy correction to account for our policy recommending multiple items at a time; (4) showcasing the value of exploration.  We demonstrate the efficacy of our approaches through a series of simulations and multiple live experiments on \youtube.   
\end{abstract}

%
%

\maketitle

{\fontsize{8pt}{8pt} \selectfont
\textbf{ACM Reference Format:}\\
Minmin Chen, Alex Beutel, Paul Covington, Sagar Jain, Francois Belletti, Ed H. Chi. 2019. Top-K Off-Policy Correction for a REINFORCE Recommender System. In \textit{The Twelfth ACM International Conference on Web Search and Data Mining (WSDM' 19), February 11-15, 2019, Melbourne, VIC, Australia}. ACM, New York, NY, USA, 9 pages. \url{https://doi.org/10.1145/3289600.3290999}
}

\section{Introduction}

Recommender systems are relied on, throughout industry, to help users sort through huge corpuses of content and discover the small fraction of content they would be interested in.  This problem is challenging because of the huge number of items that could be recommended.  Furthermore, surfacing the right item to the right user at the right time requires the recommender system to constantly adapt to users' shifting interest (state) based on their historical interaction with the system \cite{BeutelCJXLGC18}.  Unfortunately,  we observe relatively little data for such a large state and action space, with most users only having been exposed to a small fraction of items and providing explicit feedback to an even smaller fraction. That is, recommender systems receive extremely sparse data for training in general, e.g., the Netflix Prize dataset was only 0.1\% dense \cite{bennett2007netflix}. 
As a result, a good amount of research in  recommender systems explores different mechanisms for treating this extreme sparsity.  Learning from implicit user feedback, such as clicks and dwell-time, as well as filling in unobserved interactions, has been an important step in improving recommenders \cite{hu2008collaborative} but the problem remains an open one.

In a mostly separate line of research, reinforcement learning (RL) has recently achieved impressive advances in games \cite{silver2016mastering,tesauro1995temporal} as well as robotics \cite{levine2016end,kober2013reinforcement}. RL in general focuses on building agents that take actions in an environment so as to maximize some notion of long term reward. Here we explore framing recommendation as building RL agents to maximize each user's long term satisfaction with the system. This offers us new perspectives on recommendation problems as well as opportunities to build on top of the recent RL advancement. 
However, there are significant challenges to put this perspective into practice. 

As introduced above, recommender systems deal with large state and action spaces, and this is particularly exacerbated in industrial settings.
The set of items available to recommend is non-stationary and new items are brought into the system constantly, resulting in an ever-growing action space with new items having even sparser feedback.  Further, user preferences over these items are shifting all the time, resulting in continuously-evolving user states.  Being able to reason through these large number of actions in such a complex environment poses unique challenges in applying existing RL algorithms. 
Here we share our experience adapting the REINFORCE algorithm \cite{williams1992simple} to a neural candidate generator (a top-$K$ recommender system) with extremely large action and state spaces.

In addition to the massive action and state spaces, RL for recommendation is distinct in its limited availability of data.
Classic RL applications have overcome data inefficiencies by collecting large quantities of training data with self-play and simulation~\cite{silver2016mastering}.
In contrast, the complex dynamics of the recommender system has made  simulation for generating realistic recommendation data non-viable.
As a result, we cannot easily probe for reward in previously unexplored areas of the state and action space, since observing reward requires giving a real recommendation to a real user.  Instead, the model relies mostly on data made available from the previous recommendation models (policies), most of which we cannot control or can no longer control. 
To most effectively utilize logged-feedback from other policies, we take an off-policy learning approach, in which we simultaneously learn a model of the previous policies and incorporate it in correcting the data biases when training our new policy.  
We also experimentally demonstrate the value in exploratory data. 

Finally, most of the research in RL focuses on producing a policy that chooses a single item.  Real-world recommenders, on the other hand, typically offer the user multiple recommendations at a time~\cite{swaminathan2017off}.  Therefore, we define a novel top-$K$ off-policy correction for our top-$K$ recommender system.  We find that while the standard off-policy correction results in a policy that is optimal for top-1 recommendation, this top-$K$ off-policy correction leads to significant better top-$K$ recommendations in both simulations and live experiments.
Together, we offer the following contributions:
\begin{itemize}
    \item \textbf{REINFORCE Recommender:}  We scale a REINFORCE policy-gradient-based approach to learn a neural recommendation policy in a extremely large action space.
    \item \textbf{Off-Policy Candidate Generation:} We apply off-policy correction to learn from logged feedback, collected from an ensemble of prior model policies. We incorporate a learned neural model of the  behavior policies to correct data biases.
    \item \textbf{Top-$K$ Off-Policy Correction:} We offer a novel top-$K$ off-policy correction to account for the fact that our recommender outputs multiple items at a time.
    \item \textbf{Benefits in Live Experiments:} We demonstrate in live experiments, which was rarely done in existing RL literature, the value of these approaches to improve user long term satisfaction.
\end{itemize}
We find this combination of approaches valuable for increasing user enjoyment and believe it frames many of the practical challenges going forward for using RL in recommendations.





\section{Related Work}

\paragraph{Reinforcement Learning:}
Value-based approaches such as Q-learning, and policy-based ones such as policy gradients
constitute classical approaches to solve RL problems \cite{sutton1998reinforcement}.
A general comparison of modern RL approaches can be found in \cite{mnih2016asynchronous} with a focus on asynchronous learning which is key to scaling up to large problems. 
Although value-based methods present many advantages such as seamless off-policy learning, they are known to be prone to instability with function approximation~\cite{sutton2000policy}. Often, extensive hyper-parameter tuning is required to achieve stable behavior for these approaches. Despite the practical success of many value-based approaches such as deep Q-learning~\cite{mnih2013playing}, policy convergence of these algorithms are not well-studied.
Policy-based approaches on the other hand,  remain rather stable w.r.t. function approximations given a sufficiently small learning rate.
We therefore choose to rely on a policy-gradient-based approach, in particular REINFORCE \cite{williams1992simple}, and to adapt this on-policy method to provide reliable policy gradient estimates when training off-policy. 

\paragraph{Neural Recommenders:} Another line of work that is closely related to ours is the growing body of literature on applying deep neural networks to recommender systems~\cite{sedhain2015autorec,he2017neural,covington2016deep}, in particular using recurrent neural networks to incorporate temporal information and historical events for recommendation~\cite{hidasi2015session,tan2016improved,jing2017neural,WuABSJ17,BeutelCJXLGC18}. We employed similar network architectures to model the evolving of user states through interactions with the recommender system. As neural architecture design is not the main focus of our work, we refer interested readers to these prior works for more detailed discussions.

\paragraph{Bandit Problems in recommender systems:}
On-line learning methods are also popular to quickly adapt recommendation systems as new user feedback becomes available.
Bandit algorithms such as Upper Confidence Bound (UCB) \cite{auer2002finite}
trade off exploration and exploitation in an analytically tractable way that provides strong guarantees on the regret.
Different algorithms such as Thomson sampling \cite{chapelle2011empirical}, have been successfully applied to news recommendations and display advertising.
Contextual bandits offer a context-aware refinement of the basic on-line learning approaches and tailor the recommendation toward user interests~\cite{li2010contextual}.
~\citet{agarwal2014taming} aimed to make contextual bandits  tractable and easy to implement. Hybrid methods that rely on matrix factorization and bandits have also been developed to solve cold-start problems in recommender systems \cite{mary2015bandits}.

\paragraph{Propensity Scoring and Reinforcement Learning in Recommender Systems:}
The problem of learning off-policy~\cite{precup2000eligibility,precup2001off,munos2016safe} is pervasive in RL and affects policy gradient generally. As a policy evolves so does the distribution under which gradient expectations are computed.
Standard approaches in robotics \cite{schulman2015trust, achiam2017constrained} circumvent this issue by constraining policy updates so that they do not change the policy too substantially before new data is collected under an updated policy, which in return provides monotonic improvement guarantees of the RL objective. 
Such proximal methods are unfortunately not applicable in the recommendations setting where item catalogues and user behaviors change rapidly, and therefore substantial policy changes are required. Meanwhile feedback is slow to collect at scale w.r.t. the large state and action space. As a matter of fact,  offline evaluation of a given policy  is already a challenge in the recommender system setting. Multiple off-policy estimators leveraging inverse-propensity scores, capped inverse-propensity scores and various variance control measures have been developed  \cite{swaminathan2015self,thomas2016data,gilotte2018offline,swaminathan2015batch}.
Off-policy evaluation corrects for a similar data skew as off-policy RL and similar methods are applied on both problems. Inverse propensity scoring has also been employed to improve a serving policy at scale in \cite{strehl2010learning}. \citet{joachims2017unbiased} learns a model of logged feedback for an unbiased ranking model; we take a similar perspective but use a DNN to model the logged behavior policy required for the off-policy learning.
More recently an off-policy approach has been adapted to the more complex problem of slate recommendation \cite{swaminathan2017off} where a pseudo-inverse estimator assuming a structural prior on the slate and reward is applied in conjunction with inverse propensity scoring.


\section{Reinforce Recommender}

We begin with describing the setup of our recommender system, and our approach to RL-based recommendation.  





For each user, we consider a sequence of user historical interactions with the system, recording the actions taken by the recommender, \emph{i.e.,} videos recommended, as well as user feedback, such as clicks and watch time. Given such a sequence, we predict the next action to take, \emph{i.e.}, videos to recommend, so that user satisfaction metrics, \eg indicated by clicks or watch time, improve.

We translate this setup into a Markov Decision Process (MDP) 
$\left(
    \mathcal{S},
    \mathcal{A},
    \mathbf{P},
    R,
    \rho_0,
    \gamma
\right)$
where 
\begin{itemize}
\item $\mathcal{S}$: a continuous state space describing the user states;
\item $\mathcal{A}$: a discrete action space, containing items available for recommendation;
\item $\mathbf{P}: \mathcal{S} \times \mathcal{A} \times \mathcal{S} \rightarrow \mathbb{R}$ is the state transition probability; 
\item $R: \mathcal{S} \times \mathcal{A} \rightarrow \mathbb{R}$ is the reward function, where $r(s, a)$ is the immediate reward obtained by performing action $a$ at user state $s$;
\item $\rho_0$ is the initial state distribution;
\item $\gamma$ is the discount factor for future rewards.
\end{itemize}
We seek a policy $\pi(a|s)$ that casts a distribution over the item to recommend $a \in \mathcal{A}$ conditional to the user state $s \in \mathcal{S}$, 
so as to maximize the expected cumulative reward obtained by the recommender system,
$$\max_{\pi} \quad \mathcal{J}(\pi) = 
\mathbb{E}_{\tau \sim \pi}
\left[
    R(\tau) 
\right], \mbox{ where } R(\tau) = \sum_{t=0}^{|\tau|} r(s_t, a_t)
$$
Here the expectation is taken over the trajectories $\tau = (s_0, a_0, s_1, \cdots)$ obtained by acting according to the policy: $s_0\sim \rho_0, a_t\sim \pi(\cdot|s_t), s_{t+1}\sim \mathbf{P}(\cdot|s_t, a_t)$. In other words,
\begin{eqnarray}
{\mathcal{J}}(\pi) & = & \mathbb{E}_{s_0\sim \rho_0, a_t\sim \pi(\cdot|s_t), s_{t+1}\sim \mathbf{P}(\cdot|s_t, a_t)} \left[\sum_{t=0}^{|\tau|} r(s_t, a_t)\right] \nonumber\\
& = & \mathbb{E}_{s_t \sim d^\pi_t(\cdot), a_t\sim \pi(\cdot|s_t)} \left[\sum_{t'=t}^{|\tau|} r(s_{t'}, a_{t'})\right]
\end{eqnarray}
Here $d_t^\pi(\cdot)$ denotes the (discounted) state visitation frequency at time $t$ under the policy $\pi$.
Different families of methods are  available to solve such an RL problems: Q-learning \cite{silver2016mastering}, Policy Gradient \cite{williams1992simple,levine2013guided,schulman2015trust} and black box optimization \cite{hansen2001completely}.
Here we focus on a policy-gradient-based approach, \emph{i.e.,} REINFORCE \cite{williams1992simple}. 

We assume a function form of the policy $\pi_\theta$, parametrised by $\theta \in \mathbb{R}^d$. The gradient of the expected cumulative reward with respect to the policy parameters can be derived analytically thanks to the ``log-trick'', yielding the following REINFORCE gradient
\begin{eqnarray}
\label{eq:reinforce}
\nabla_\theta \mathcal{J}(\pi_\theta)  &=& \mathbb{E}_{s_t \sim d^\pi_t(\cdot), a_t\sim \pi(\cdot|s_t)}
\left[
     \left(\sum_{t'=t}^{|\tau|} r(s_{t'}, a_{t'})\right)\nabla_{\theta} \log \pi_\theta(a_t|s_t)
\right]\nonumber\\
&=& \sum_{s_t \sim d^\pi_t(\cdot), a_t\sim \pi(\cdot|s_t)}
     R_t(s_t, a_t)\nabla_{\theta} \log \pi_\theta(a_t|s_t)
\end{eqnarray}
Here $R_t(s_t, a_t) = \sum_{t'=t}^{|\tau|} \gamma^{t'-t} r(s_{t'}, a_{t'})$ is the discounted future reward for action at time $t$. The discounting factor $\gamma$ is applied to reduce variance in the gradient estimate.
In on-line RL, where the policy gradient is computed on trajectories generated by the policy under consideration, the monte carlo estimate of the policy gradient is unbiased.


\section{Off-Policy Correction}
Unlike classical reinforcement learning, our learner does not have real-time interactive control of the recommender due to learning and infrastructure constraints. In other words, we cannot perform online updates to the policy and generate trajectories according to the updated policy immediately. Instead we receive logged feedback of actions chosen by a historical policy (or a mixture of policies), which could have a different distribution over the action space than the policy we are updating. 

We focus on addressing the data biases that arise when applying policy gradient methods under this setting. In particular, the fact that we collect data with a periodicity of several hours and compute many policy parameter updates before deploying a new version of the policy in production implies that the set of trajectories we employ to estimate the policy gradient is generated by a different policy. Moreover, we learn from batched feedback collected by other recommenders as well, which follow drastically different policies. A naive policy gradient estimator is no longer unbiased as the gradient in \eq~\eqref{eq:reinforce} requires sampling trajectories from the updated policy $\pi_\theta$ while the trajectories we collected were drawn from a combination of historical policies $\beta$. 


We address the distribution mismatch with importance weighting~\cite{precup2000eligibility,precup2001off,munos2016safe}.
Consider a trajectory $\tau = (s_0, a_0, s_1, \cdots)$ sampled according to a behavior policy $\beta$,
the off-policy-corrected gradient estimator is then:
\begin{equation}
\nabla_\theta \mathcal{J}(\pi_\theta) = \sum_{s_t \sim d^\beta_t(\cdot), a_t\sim \beta(\cdot|s_t)}
\omega(s_t, a_t) R_t\nabla_{\theta} \log \pi_\theta(\tau)
\end{equation}
where
$$\omega(s_t, a_t) = \frac{d^\pi_t(s_t)}{d^\beta_t(s_t)}\times \frac{\pi_\theta(a_t|s_t)}{\beta(a_t|s_t)} \times \prod_{t'=t+1}^{|\tau|}\frac{\pi_\theta(a_{t'}|s_{t'})}{\beta_\theta(a_{t'}|s_{t'})}$$
is the importance weight. This correction produces an unbiased estimator whenever the trajectories
are collected with actions sampled according to $\beta$.
However, the variance of the estimator can be huge when the difference in $\pi_\theta$ and the behavior policy $\beta$ results in very low or high values of the importance weights.

To reduce the variance of each gradient term, we take the first-order approximation and ignore the state visitation differences under the two policies as the importance weights of future trajectories, 
which yields a slightly biased estimator of the policy gradient with lower variance:
\begin{equation}~\label{eq:offpolicy_reinforce}
\nabla_\theta \mathcal{J}(\pi_\theta) \approx \sum_{s_t \sim d^\beta_t(\cdot), a_t\sim \beta(\cdot|s_t)}
\frac{\pi_\theta(a_t|s_t)}{\beta(a_t|s_t)} R_t\nabla_{\theta} \log \pi_\theta(\tau)
\end{equation}
\citet{achiam2017constrained} prove that the impact of this first-order approximation on the total reward of the learned policy is bounded in magnitude by
$
O
\left(
E_{s \sim d^\beta}
\left[
    D_{TV}(\pi | \beta)[s]
\right]
\right)
$
where $D_{TV}$ is the total variation between $\pi(\cdot|s)$ and $\beta(\cdot|s)$ and $d^{\beta}$ is the discounted future state distribution under $\beta$. 
This estimator trades off the variance of the exact off-policy correction while still correcting for the large bias of a non-corrected policy gradient, which is better suited for on-policy learning.

\subsection{Parametrising the policy $\pi_\theta$}
We model our belief on the user state  at each time $t$, which capture both evolving user interests using a $n$-dimensional vector, that is, $\mathbf{s}_t \in \mathbb{R}^n$. The action taken at each time $t$ along the trajectory is embedded using an $m$-dimensional vector $\mathbf{u}_{a_t} \in \mathbb{R}^m$. 
We model the state transition $\mathbf{P}: \mathcal{S}\times\mathcal{A}\times\mathcal{S}$ with a recurrent neural network~\cite{WuABSJ17,BeutelCJXLGC18}:
$$\mathbf{s}_{t+1} = f(\mathbf{s}_t, \mathbf{u}_{a_t}).$$
We experimented with a variety of popular RNN cells such as Long Short-Term Memory (LSTM)~\cite{hochreiter1997long} and Gated Recurrent Units (GRU)~\cite{chung2014empirical}, and ended up using a simplified cell called Chaos Free RNN (CFN)~\cite{laurent2016recurrent} due to its stability and computational efficiency. The state is updated recursively as 
\begin{align}\label{eq:cfn}
\mathbf{s}_{t+1} & = \mathbf{z}_t \odot \tanh(\mathbf{s}_{t}) + \mathbf{i}_t \odot \tanh(\mathbf{W}_a \mathbf{u}_{a_t})\\
\mathbf{z}_t & = \sigma(\mathbf{U}_z \mathbf{s}_t + \mathbf{W}_z \mathbf{u}_{a_t} + \mathbf{b}_z)\nonumber\\
\mathbf{i}_t & = \sigma(\mathbf{U}_i \mathbf{s}_t + \mathbf{W}_i \mathbf{u}_{a_t} + \mathbf{b}_i)\nonumber
\end{align}
where
$\mathbf{z}_t, \mathbf{i}_t \in \mathbb{R}^n$ are the update and input gate respectively. 

Conditioning on a user state $\mathbf{s}$, the policy $\pi_\theta(a | \mathbf{s})$ is then modeled with a simple softmax, 
\begin{equation}\label{eq:softmax}
    \pi_\theta(a | \mathbf{s}) = \frac{exp(\mathbf{s}^\top \mathbf{v}_{a}/T)}{\sum_{a'\in\mathcal{A}} exp(\mathbf{s}^\top \mathbf{v}_{a'}/T)}
\end{equation}
where $\mathbf{v}_a \in \mathbb{R}^n$ is another embedding for each action $a$ in the action space $\mathcal{A}$ and $T$ is a temperature that is normally set to 1. Using a higher value in $T$ produces a smoother policy over the action space. The normalization term in the softmax requires going over all the possible actions, which is in the order of millions in our setting. To speed up the computation, we perform sampled softmax~\cite{bengio2003quick} during training. At serving time, we used an efficient nearest neighbor search algorithm to retrieve top actions and approximate the softmax probability using these actions only, as detailed in section~\ref{sec:exploration}. 

\begin{figure}[h]
\centering
\includegraphics[width=0.45\textwidth]{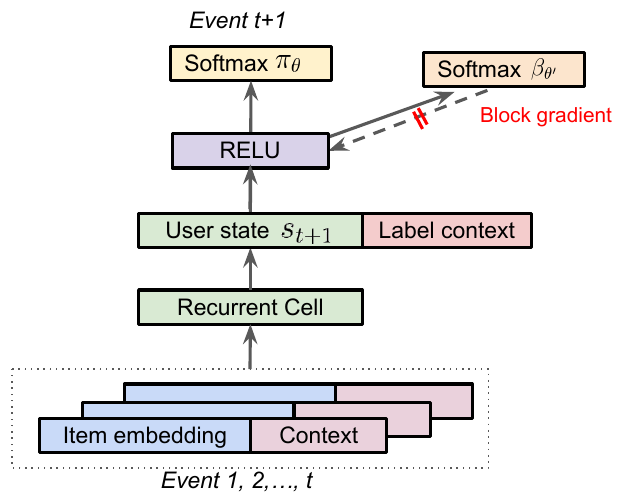} 
  \caption{A diagram shows the parametrisation of the policy $\pi_\theta$ as well as the behavior policy $\beta_{\theta'}$.}
  \label{fig:model_architecture}
\end{figure}

In summary, the parameter $\theta$ of the policy  $\pi_\theta$ contains the two action embeddings $\mathbf{U} \in \mathbb{R}^{m\times |\mathcal{A}|}$ and $\mathbf{V} \in \mathbb{R}^{n\times|\mathcal{A}|}$ as well as the weight matrices $\mathbf{U}_z, \mathbf{U}_i \in \mathbb{R}^{n\times n}$, $\mathbf{W}_u, \mathbf{W}_i, \mathbf{W}_a \in \mathbb{R}^{n\times m}$ and biases $\mathbf{b}_u, \mathbf{b}_i \in \mathbb{R}^n$ in the RNN cell.   Figure~\ref{fig:model_architecture} shows a diagram describing the neural architecture of the main policy  $\pi_\theta$.  Given an observed trajectory $\tau = (s_0, a_0, s_1, \cdots)$
sampled from a behavior policy $\beta$, the new policy first generates a model of the user state $\mathbf{s}_{t+1}$ by starting with an initial state $\mathbf{s}_0 \sim \rho_0$\footnote{In our experiment, we used a fixed initial state distribution, where $\mathbf{s}_0 = \mathbf{0} \in \mathbb{R}^n$} and iterating through the recurrent cell as in \eq~\eqref{eq:cfn}\footnote{We take into account of the context~\cite{BeutelCJXLGC18} of the action, such as page, device and time information, as input to the RNN cell besides the action embedding itself.}. Given the user state $\mathbf{s}_{t+1}$ the policy head casts a distribution on the action space through a softmax as in \eq \eqref{eq:softmax}. With $\pi_\theta(a_{t+1}|\mathbf{s}_{t+1})$, we can then produce a policy gradient as in \eq \eqref{eq:offpolicy_reinforce} to update the policy.

\subsection{Estimating the behavior policy $\beta$}

One  difficulty in coming up with the off-policy corrected estimator in \eq \eqref{eq:offpolicy_reinforce} is to get the behavior policy $\beta$. Ideally, for each logged feedback of a chosen action we received, we would like to also log the probability of the behavior policy choosing that action. Directly logging the behavior policy is however not feasible in our case as (1) there are multiple agents in our system, many of which we do not have control over, and (2) some agents have a deterministic policy, and setting $\beta$ to 0 or 1 is not the most effective way to utilize these logged feedback. 

Instead we take the approach first introduced in~\citep{strehl2010learning}, and estimate the behavior policy $\beta$, which in our case is a mixture of the policies of the multiple agents in the system, using the logged actions. Given a set of logged feedback $\mathcal{D} = \{(\mathbf{s}_i, a_i), i = 1, \cdots, N\}$, ~\citet{strehl2010learning} estimates $\hat\beta(a)$ independent of user state by aggregate action frequency throughout the corpus.  In contrast, we adopt a context-dependent neural estimator. For each state-action pair $(s, a)$ collected, we estimate the probability $\hat\beta_{\theta'}(a|s)$ that the mixture of behavior policies choosing that action using another softmax, parametrised by $\theta'$. As shown in Figure~\ref{fig:model_architecture}, we re-use the user state $s$ generated from the RNN model from the main policy, and model the mixed behavior policy with another softmax layer. To prevent the behavior head from intefering with the user state of the main policy, we block its gradient from flowing back into the RNN. We also experimented with separating the $\pi_\theta$ and $\beta_{\theta'}$ estimators, which incurs computational overhead for computing another state representation but does not results in any metric improvement in offline and live experiments. 

Despite a substantial sharing of parameters between the two policy heads $\pi_\theta$ and $\beta_{\theta'}$, there are two noticeable difference between them: (1) While the main policy $\pi_\theta$ is effectively trained using a weighted softmax to take into account of long term reward, the behavior policy head $\beta_{\theta'}$ is trained using only the state-action pairs; (2) While the main policy head $\pi_\theta$ is trained using only items on the trajectory with non-zero reward  
~\footnote{1. Actions with zero-reward will not contribute to the gradient update in $\pi_\theta$; 2. We ignore them in the user state update as users are unlikely to notice them and as a result, we assume the user state are not influenced by these actions; 3. It saves computational cost.}, the behavior policy  $\beta_{\theta'}$ is trained using all of the items on the trajectory to avoid introducing bias in the $\beta$ estimate. 


In~\citep{strehl2010learning}, it is argued that that a behavior policy that is deterministically choosing an action $a$ given state $s$ at time $t_1$ and action $b$ at time $t_2$ can be treated as randomizing between action $a$ and $b$ over the timespan of the logging. Here we could argue the same point, which explains why the behavior policy could be other than 0 or 1 given a deterministic policy. In addition, since we have multiple policies acting simultaneously, if one policy is determinstically choosing action $a$ given user state $s$, and another one is determinstically choosing action $b$, then estimating $\hat\beta_{\theta'}$ in such a way would approximate the expected frequency of action $a$ being chosen under the mixture of these behavior policies given user state $s$.

\subsection{Top-$K$ Off-Policy Correction}~\label{sec:topk_offpolicy}
Another challenge in our setting is that our system recommends a page of $k$ items to users at a time. As users are going to browse through (the full or partial set of) our recommendations and potentially interact with more than one item, we need to pick a set of relevant items instead of a single one. In other words, we seek a policy $\Pi_\theta(A|s)$, here each action $A$ is to select a set of $k$ items, to maximize the expected cumulative reward,
$$\max_\theta \mathcal{J}(\Pi_\theta) =  \mathbb{E}_{s_t\sim d^\Pi_t(\cdot), A_t \sim \Pi_\theta(\cdot|s_t)}\left[R_t(s_t, A_t)\right].$$
Here $R_t(s_t, A_t)$ denotes the cumulative return of the set $A_t$ at state $s_t$.
Unfortunately, the action space grows exponentially under this set recommendation formulation~\cite{swaminathan2017off,zhao2018deep}, which is prohibitively large given the number of items we choose from are in the orders of millions.  

To make the problem tractable, we assume that a user will interact with \textbf{at most} one item from the returned set $A$. In other words, there will be at most one item with non-zero cumulative reward among $A$. We further assume that the expected return of an item is independent of other items chosen in the set $A$~\footnote{This assumption holds if the downstream systems and the user inspect each item on a page independently.}. With these two assumptions, we can reduce the set problem to 
$$\mathcal{J}(\Pi_\theta) =  \mathbb{E}_{s_t\sim d^\Pi_t(\cdot), a_t \in A_t \sim \Pi_\theta(\cdot|s_t)}\left[R_t(s_t, a_t)\right].$$
Here $R_t(s_t, a_t)$ is the cumulative return of the item $a_t$ the user interacted with, and $a_t\in A_t \sim \Pi_\theta(\cdot|s_t)$ indicates that $a_t$ was chosen by the set policy. 
Furthermore, we constrain ourselves to generate the set action $A$ by independently sampling each item $a$ according to the softmax policy $\pi_\theta$ described in  \eq~\eqref{eq:softmax} and then de-duplicate. As a result, the probability of an item $a$ appearing in the final non-repetitive set $A$ is simply $\alpha_\theta(a|s) = 1 - (1-\pi_\theta(a|s))^{K}$, where $K$ is the number of times we sample.~\footnote{At a result of the sampling with replacement and de-duplicate, the size of the final set $A$ can vary.}. 

We can then adapt the REINFORCE algorithm to the set recommendation setting by simply modifying the gradient update in \eq~\eqref{eq:reinforce} to
$$ \sum_{s_t \sim d^\pi_t(\cdot), a_t\sim \alpha_\theta(\cdot|s_t)}
     R_t(s_t, a_t)\nabla_{\theta} \log \alpha_\theta(a_t|s_t)$$

Accordingly, we can update the off-policy corrected gradient in \eq~\eqref{eq:offpolicy_reinforce} by replacing $\pi_\theta$ with $\alpha_\theta$, resulting in the top-$K$ off-policy correction factor:
\begin{align}~\label{eq:topk_reinforce}
&\sum_{s_t \sim d^\pi_t(\cdot), a_t\sim \beta(\cdot|s_t)}\left[
    \frac{
            \alpha_\theta \left(a_{t} | s_{t} \right)
        }{
            \beta \left(a_{t} | s_{t} \right)
    }
    R_t(s_t, a_t) \nabla_{\theta} \log
        \alpha_\theta (a_t | s_t)
\right]\\
 = &\sum_{s_t \sim d^\pi_t(\cdot), a_t\sim \beta(\cdot|s_t)}\left[
    \frac{
            \pi_\theta \left(a_{t} | s_{t} \right)
        }{
            \beta \left(a_{t} | s_{t} \right)
    }
    \frac{
            \partial \alpha \left(a_{t} | s_{t} \right)
        }{
            \partial \pi \left(a_{t} | s_{t} \right)
    }
    R_t(s_t, a_t) \nabla_{\theta} \log
        \pi_\theta (a_t | s_t)
\right].\nonumber
\end{align}
Comparing \eq~\eqref{eq:topk_reinforce} with \eq~\eqref{eq:offpolicy_reinforce}, the top-$K$ policy adds an additional multiplier of
\begin{align}
    \lambda_K(s_t, a_t) = \frac{
            \partial \alpha \left(a_{t} | s_{t} \right)
        }{
            \partial \pi \left(a_{t} | s_{t} \right)
    } =  K(1 - \pi_\theta(a_t | s_t))^{K-1}  
\end{align}
to the original off-policy correction factor of $\frac{\pi(a|s)}{\beta(a|s)}$. 

Now let us take a closer look at this additional multiplier:
\begin{itemize}
\item As $\pi_\theta(a |s) \rightarrow 0$, $\lambda_K(s, a)\rightarrow K$. The top-$K$ off-policy correction increases the policy update by a factor of $K$ comparing to the standard off-policy correction;
\item As $\pi_\theta(a |s) \rightarrow 1$, $\lambda_K(s, a)\rightarrow 0$. This multiplier zeros out the policy update.
\item As $K$ increases, this multiplier reduces the gradient to zero faster as $\pi_\theta(a |s)$ reaches a reasonable range. 
\end{itemize}
In summary, when the desirable item has a small mass in the softmax policy $\pi_\theta(\cdot|s)$, the top-$K$ correction more aggressively pushes up its likelihood than the standard correction. Once the softmax policy $\pi_\theta(\cdot|s)$ casts a reasonable mass on the desirable item (to ensure it will be likely to appear in the top-$K$), the correction then zeros out the gradient and no longer tries to push up its likelihood. This in return allows other items of interest to take up some mass in the softmax policy. As we are going to demonstrate in the simulation as well as live experiment, while the standard off-policy correction converges to a policy that is optimal when choosing a single item, the top-$K$ correction leads to better top-$K$ recommendations.

\subsection{Variance Reduction Techniques}~\label{sec:variance_reduction}
As detailed at the beginning of this section, we take a first-order approximation to reduce variance in the gradient estimate. Nonetheless, the gradient can still suffer from large variance due to large importance weight of $\omega(s, a) = \frac{\pi(a|s)}{\beta(a|s)}$ as shown in \eq \eqref{eq:offpolicy_reinforce}, 
Similarly for top-$K$ off-policy correction.
Large importance weight could result from (1) large deviation of the new policy $\pi(\cdot|s)$ from the behavior policy, in particular, the new policy  explores regions that are less explored by the behavior policy. That is, $\pi(a|s) \gg \beta(a|s)$ and (2) large variance in the $\beta$ estimate.

We tested several techniques proposed in counterfactual learning and RL literature to control variance in the gradient estimate. Most of these techniques reduce variance at the cost of introducing some bias in the gradient estimate. 

\textbf{Weight Capping.} The first approach we take is to simply cap the weight~\cite{bottou2013counterfactual} as 
\begin{equation}
\bar{\omega}_c(s, a) = \min\left(\frac{\pi(a|s)}{\beta(a|s)}, c\right).
\label{eq:capping}
\end{equation}
Smaller value of $c$ reduces variance in the gradient estimate, but introduces larger bias.

\textbf{Normalized Importance Sampling (NIS).} Second technique we employed is to introduce a ratio control variate, where we use classical weight normalization~\cite{mcbook} defined by:
$$\bar{\omega}_n(s, a) = \frac{\omega(s, a)}{\sum_{(s', a')\sim \beta}\omega(s', a')}.$$
As $\mathbb{E}_\beta[\omega(s, a)] = 1$, the normalizing constant is equal to $n$, the batch size, in expectation. As $n$ increases, the effect of NIS is equivalent to tuning down the learning rate.

\textbf{Trusted Region Policy Optimization (TRPO).} TRPO~\cite{schulman2015trust} prevents the new policy $\pi$ from deviating from the behavior policy by adding a regularization that penalizes the KL divergence of these two policies. It achieves similar effect as the weight capping.



\section{Exploration}~\label{sec:exploration}
As should be clear by this point, the distribution of training data is important for learning a good policy.  
Exploration policies to inquire about actions rarely taken by the existing system have been extensively studied. 
In practice, brute-force exploration, such as $\epsilon$-greedy, is not viable in a production system like YouTube where this could, and mostly likely would, result in inappropriate recommendations and a bad user experience.  For example, \citet{schnabel2018short} studied the cost of exploration.  

Instead we employ Boltzmann exploration~\cite{daw2006cortical} to get the benefit of exploratory data without negatively impacting user experience. We consider using a stochastic policy where recommendations are sampled from $\pi_\theta$ rather than taking the $K$ items with the highest probability.  
This has the challenge of being computationally inefficient because we need to calculate the full softmax, which is prohibitively expensive considering our action space.  Rather, we make use of efficient approximate nearest neighbor-based systems to look up the top $M$ items in the softmax \cite{guo2016quantization}.  We then feed the logits of these $M$ items into a smaller softmax to normalize the probabilities and sample from this distribution.  By setting $M \gg K$ we can still retrieve most of the probability mass, limit the risk of bad recommendations, and enable computationally efficient sampling.
In practice, we further balance exploration and exploitation by returning the top $K'$ most probable items and sample $K-K'$ items from the remaining $M-K'$ items.


\section{Experimental Results}
We showcase the effectiveness of these approaches for addressing data biases in a series of simulated experiments and live experiments in an industrial-scale recommender system. 

\subsection{Simulation}

\begin{figure}[t!]
    \centering
\includegraphics[width=0.23\textwidth]{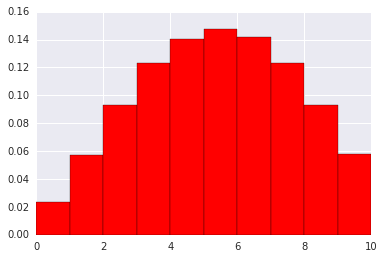} 
\includegraphics[width=0.23\textwidth]{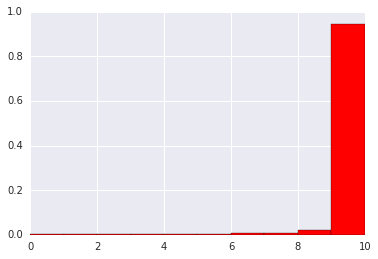} 
    \caption{learned policy $\pi_\theta$ when behavior policy $\beta$ is skewed to favor the actions with least reward, \emph{i.e.}, $\beta(a_i) = \frac{11-i}{55}, \forall i = 1, \cdots, 10$. (left): without off-policy correction; (right): with off-policy correction.}
    \label{fig:simulation1}
\vspace{-0.1in}
\end{figure}

We start with designing  simulation experiments to shed light on the off-policy correction ideas under more controlled settings. To simplify our simulation, we assume the problem is stateless, in other words, the reward $R$ is independent of user states, and the action does not alter the user states either. As a result, each action on a trajectory can be independently chosen. 

\subsubsection{Off-policy correction}
 In the first simulation, we assume there are 10 items, that is $\mathcal{A} = \{a_i, i = 1, \cdots, 10\}$. The reward of each one is equal to its index, that is, $r(a_i) = i$. When we are choosing a single item, the optimal policy under this setting is to always choose the $10^{th}$ item as it gives the most reward, that is,
$$\pi^\ast(a_i) = \mathbb{I}(i=10).$$ 

We parameterize $\pi_\theta$  using a stateless softmax 
$$\pi(a_i) = \frac{e^{\theta_i}}{\sum_j e^{\theta_{j}}}$$
Given observations sampled from the behavior policy $\beta$,  naively applying policy gradient without taking into account of data bias as in \eq \eqref{eq:reinforce} would converge to a policy 
$$\pi(a_i) = \frac{r(a_i) \beta(a_i)}{\sum_{j} r(a_j) \beta(a_j)}$$
This has an obvious downside: the more the behavior policy chooses a sub-optimal item, the more the new policy will be biased toward choosing the same item. 

Figure~\ref{fig:simulation1} compares the policies $\pi_\theta$, learned without and with off-policy correction using SGD~\cite{bottou2010large}, when the behavior policy $\beta$ is skewed to favor items with least reward. 
As shown in Figure~\ref{fig:simulation1}~(left), naively applying the policy gradient without accounting for the data biases leads to a sub-optimal policy. In the worst case, if the behavior policy always chooses the action with the lowest reward, we will end up with a policy that is arbitrarily poor and mimicking the behavior policy (\emph{i.e.}, converge to selecting the least rewarded item). On the other hand, applying the off-policy correction allows us to converge to the optimal policy $\pi^\ast$ regardless of how the data is collected, as shown in Figure~\ref{fig:simulation1}~(right). 

\subsubsection{Top-$K$ off-policy correction.} \label{sec:simulation_topk_offpolicy}
To understand the difference between the standard off-policy correction and the top-$K$ off-policy correction proposed, we designed another simulation in which we can recommend multiple items. Again we assume there are 10 items, with $r(a_1) = 10, r(a_2) = 9$, and the remaining items are of much lower reward $r(a_i) = 1, \forall i = 3, \cdots, 10$. Here we focus on recommending two items, that is, $K=2$. The behavior policy $\beta$ follows a uniform distribution, \textit{i.e.,} choosing each item with equal chance. 

Given an observation $(a_i, r_i)$ sampled from $\beta$, the standard off-policy correction has a SGD updates of the following form,
$$\theta_j = \theta_j + \eta \frac{\pi_\theta(a_j)}{\beta(a_j)}r(a_i) \left[\mathbb{I}(j = i) - \pi_\theta(a_j)\right], \quad \forall j = 1, \cdots, 10 $$
where $\eta$ is the learning rate. SGD keeps increasing the likelihood of the item $a_i$  proportional to the expected reward under $\pi_\theta$ until $\pi_\theta(a_i) = 1$, under which the gradient goes to 0. The top-$K$ off-policy correction, on the other hand, has an update of the following form,
$$\theta_j = \theta_j + \eta \lambda_K(a_i)\frac{\pi_\theta(a_j)}{\beta(a_j)}r(a_i) \left[\mathbb{I}(j = i) - \pi_\theta(a_j)\right], \quad \forall j = 1, \cdots, 10 $$
where $\lambda_K(a_i)$ is the multiplier as defined in section~\ref{sec:topk_offpolicy}. When $\pi_\theta(a_i)$ is small, $\lambda_K(a_i)\approx K$, and SGD increases the likelihood of the item $a_i$ more aggressively. 
As $\pi_\theta(a_i)$ reaches to a large enough value, $\lambda_K(a_i)$ goes to 0. As a result, SGD will no longer force to increase the likelihood of this item \textbf{even when $\pi_\theta(a_i)$ is still less than 1}. This in return allows the second-best item to take up some mass in the learned policy. 

\begin{figure}[btp]
    \centering
\includegraphics[width=0.23\textwidth]{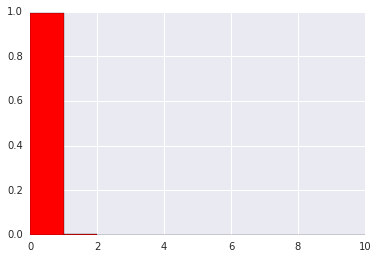} 
\includegraphics[width=0.23\textwidth]{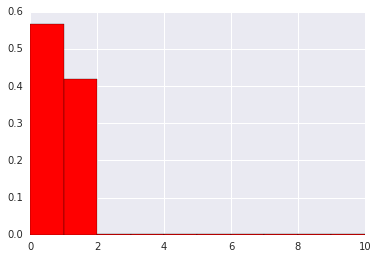} 
\caption{Learned policy $\pi_\theta$ (left): with standard off-policy correction; (right): with top-k correction for top-2 recommendation.}
    \label{fig:simulation2}
\vspace{-0.1in}
\end{figure}

Figure~\ref{fig:simulation2} shows the policies $\pi_\theta$ learned with the standard (left) and top-$K$ off-policy correction (right).  We can see that with the standard off-policy correction, although the learned policy is  calibrated~\cite{lapin2016loss} in the sense that it still maintains the ordering of items w.r.t. their expected reward, it converges to a policy that cast almost its entire mass on the top-1 item, that is $\pi(a_1) \approx 1.0$. As a result, the learned policy loses track of the difference between a slightly sub-optimal item ($a_2$ in this example) and the rest. 
The top-$K$ correction, on the other hand, converges to a policy that has a significant mass on the second optimal item, while maintaining the order of optimality between items. As a result, we are able to recommend to users two high-reward items and aggregate more reward overall.


\subsection{Live Experiments}
While simulated experiments are valuable to understand new methods, the goal of any recommender systems is ultimately to improve \emph{real} user experience.  
We therefore conduct a series of A/B experiments running in a live system to measure the benefits of these approaches.  

We evaluate these methods on a production RNN candidate generation model in use at YouTube, similar to the setup described in \cite{covington2016deep,BeutelCJXLGC18}.  The model is one of many candidate generators that produce recommendations, which are scored and ranked by a separate ranking model before being shown to users on the YouTube Homepage or the side panel on the video watch page.
As described above, the model is trained following the REINFORCE algorithm.  The immediate reward $r$ is designed to reflect different user activities;  videos that are recommended but not clicked receive zero reward. The long term reward $R$ is aggregated over a time horizon of 4--10 hours. In each experiment both the control and the test model use the same reward function. Experiments are run for multiple days, during which the model is trained continuously with new events being used as training data with a lag under 24 hours. While we look at various online metrics with the recommender system during live experiments, we are going to focus our discussion on the amount of time user spent watching videos, referred to as ViewTime. 

The experiments presented here describe multiple sequential improvements to the production system.  Unfortunately, in such a setting, the latest recommender system provides the training data for the next experiment, and as a result, once the production system incorporates a new approach, subsequent experiments cannot be compared to the earlier system.  Therefore, each of the following experiments should be taken as the analysis for each component individually, and we state in each section what was the previous recommender system from which the new approach receives data.

\subsubsection{Exploration}
We begin with understanding the value of exploratory data in improving model quality.  In particular, we would like to measure if serving a stochastic policy, under which we sample from the softmax model as described in Section \ref{sec:exploration}, results in better recommendations than serving a deterministic policy where the model always recommends the $K$ items with the highest probability according to the softmax.


We conducted a first set of experiments to understand the impact of serving a stochastic policy vs. a deterministic one while keeping the training process unchanged. In the experiment, the control population is served with a deterministic policy, while a small slice of test traffic is served with the stochastic policy as described in Section~\ref{sec:exploration}. Both policies are based on the same softmax model trained as in \eq \eqref{eq:rl_loss}. To control the amount of randomness in the stochastic policy at serving, we varied the temperature used in \eq~\eqref{eq:softmax}. A lower $T$ reduces the stochastic policy to a deterministic one, while a higher $T$ leads to a random policy that recommends any item with equal chance. With $T$ set to 1, we observed \emph{no statistically significant change} in ViewTime during the experiment, which suggests the amount of randomness introduced from sampling does not hurt the user experience directly.


However, this experimental setup does not account for the benefit of having exploratory data available during training. One of the main biases in learning from logged data is that the model does not observe  feedback of actions not chosen by the previous recommendation policy, and exploratory data alleviates this problem.  
We conducted a followup experiment where we introduce the exploratory data into training. To do that, we split users on the platform into three buckets: 90\%, 5\%, 5\%. The first two buckets are served with a deterministic policy based on a deterministic model and the last bucket of users is served with a stochastic policy based on a model trained with exploratory data. The deterministic model is trained using only data acquired in the first two buckets, while the stochastic model is trained on data from the first and third buckets. As a result, these two models receive the same amount of training data, but the stochastic model is more likely to observe the outcomes of some rarer state, action pairs because of exploration.

Following this experimental procedure, we observe a statistically significant increase in ViewTime 
by 0.07\% in the test population.  While the improvement is not large, it comes from a relatively small amount of exploration data (only 5\% of users experience the stochastic policy).  
We expect higher gain now that the stochastic policy has been fully launched.

\subsubsection{Off-Policy Correction}
\label{sec:eval_offpolicy}
Following the use of a stochastic policy, we tested incorporating off-policy correction during training. Here, we follow a more traditional A/B testing setup~\footnote{In practice, we use a fairly small portion of users as test population; as a result, the feedback we logged are almost entirely acquired by the control model.} where we train two models, both using the full traffic.  The control model is trained following \eq~\eqref{eq:rl_loss}, only weighting examples by the reward.  The test model follows the structure in Figure \ref{fig:model_architecture}, where the model learns both a serving policy $\pi_\theta$ as well as the behavior policy $\beta_{\theta'}$.  The serving policy is trained with the off-policy correction described in \eq~\eqref{eq:offpolicy_reinforce} where each example is weighted not only by the reward but also the importance weight $\frac{\pi_\theta}{\beta_{\theta'}}$ for addressing data bias.


\begin{figure}
    \centering
    \includegraphics[width=\columnwidth]{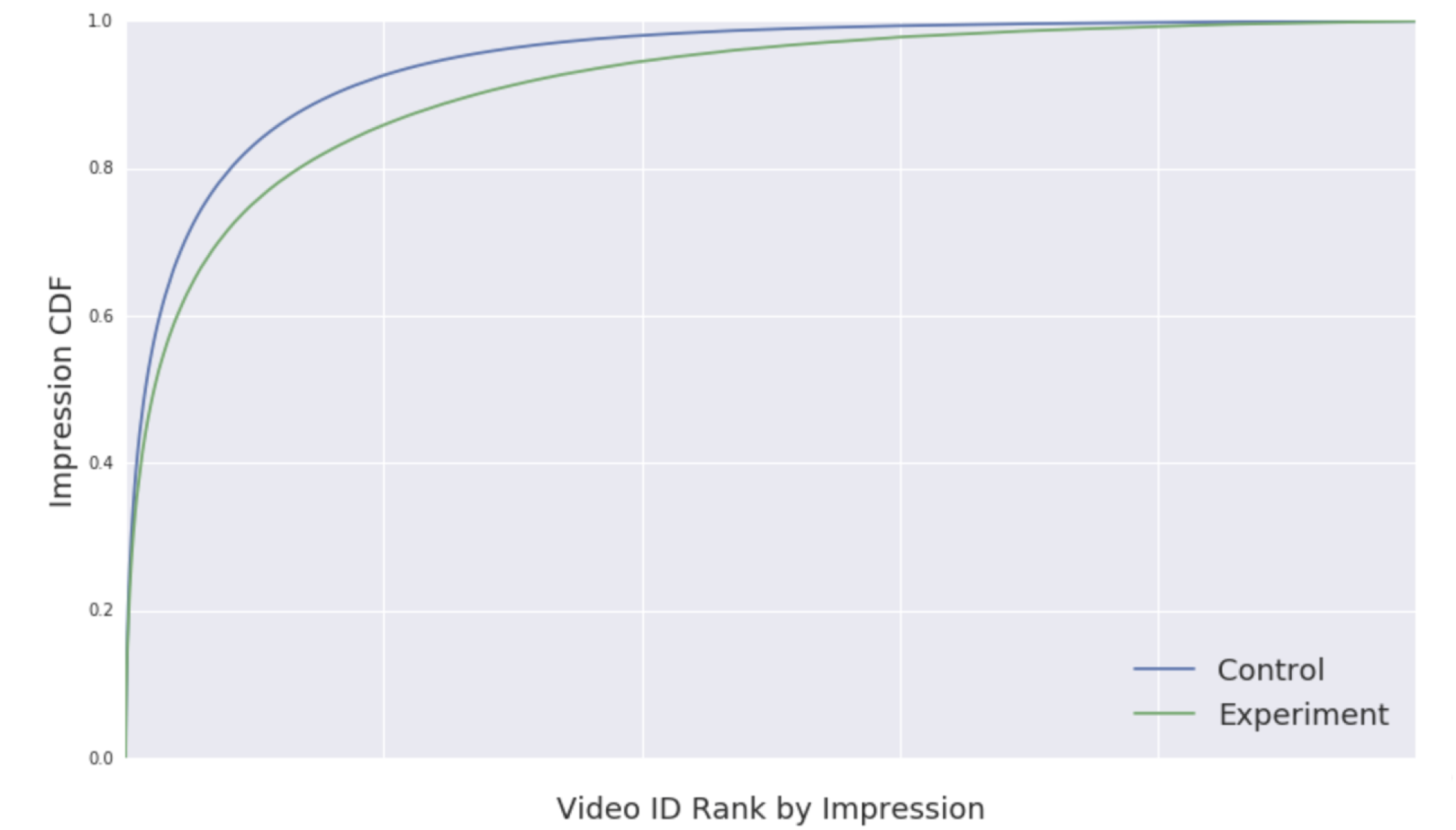}
    \caption{CDF of videos nominated in control and test population according to rank of videos in the control population. Standard off-policy correction addresses the ``rich get richer`` phenomenon.}
    \label{fig:impression_rank}
\vspace{-0.1in}
\end{figure}

During experiments, we observed the learned policy (test) starts to deviate from the behavior policy (control) that is used to acquire the traffic. Figure~\ref{fig:impression_rank} plots a CDF of videos selected by our nominator in control and experiment according to the rank of videos in control population (rank 1 is the most nominated video by the control model, and the rightmost is least nominated). We see that instead of mimicking the model (shown in blue) used for data collection, the test model (shown in green) favors videos that are less explored by the control model. We observed that the proportion of nominations coming from videos outside of the top ranks is increased by nearly a factor of three in experiment. This aligns with what we observed in the simulation shown in Figure~\ref{fig:simulation1}. While ignoring the bias in the data collection process creates a ``rich get richer`` phenomenon, whereby a video is nominated in the learned policy simply because it was heavily nominated in the behavior policy, incorporating the off-policy correction reduces this effect.  



Interestingly, in live experiment, we did not observe a statistically significant change in ViewTime between control and test population. However, we saw an increase in the number of videos viewed by 0.53\%, which was statistically significant, suggesting that users are indeed getting more enjoyment.

\subsubsection{Top-$K$ Off-Policy}
\label{sec:eval_topk}
We now focus on understanding if the top-$K$ off-policy learning improves the user experience over the standard off-policy approach. 
In this case, we launched an equivalently structured model now trained with the top-$K$ off-policy corrected gradient update given in \eq \eqref{eq:topk_reinforce} and compared its performance to the previous off-policy model, described in Section \ref{sec:eval_offpolicy}.
In this experiment, we use $K=16$ and capping $c=e^3$ in \eq \eqref{eq:capping}; we will explore these hyperparameters in more detail below.

As described in Section~\ref{sec:topk_offpolicy} and demonstrated in the simulation in Section~\ref{sec:simulation_topk_offpolicy}, while the standard off-policy correction we tested before leads to a policy that is overly-focused on getting the top-1 item correct, the top-$K$ off-policy correction converges to a smoother policy under which there is a non-zero mass on the other items of interest to users as well. 
This in turn leads to better top-$K$ recommendation. Given that we can recommend multiple items, the top-$K$ off-policy correction leads us to present a better full-page experience to users than the standard off-policy correction. 
In particular, we find that the amount of ViewTime increased by 0.85\% in the test traffic, with the number of videos viewed slightly decreasing by 0.16\%.


\subsubsection{Understanding Hyperparameters}
Last, we perform a direct comparison of how different hyperparameter choices affect the top-$K$ off-policy correction, and in turn the user experience on the platform.  We perform these tests after the top-$K$ off-policy correction became the production model.  

\begin{figure}
    \centering
    \includegraphics[width=0.9\columnwidth]{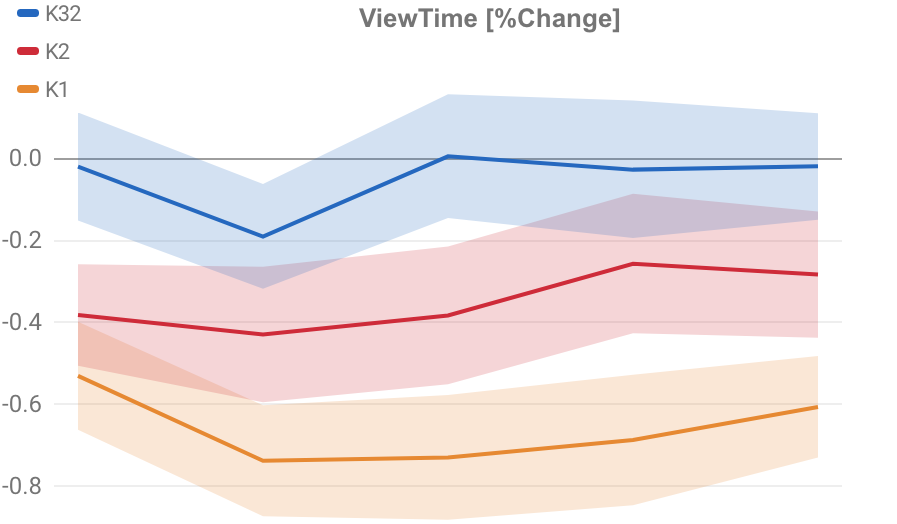}
    \caption{top-$K$ off-policy correction with varying $K$.}
    \label{fig:num_actions}
\vspace{-0.1in}
\end{figure}

\paragraph{Number of actions} We first explore the choice of $K$ in the top-$K$ off-policy correction.  We train three structurally identical models, using $K \in \{1, 2, 16, 32\}$; 
The control (production) model is the top-$K$ off-policy model with $K=16$. 
We plot the results during a 5-day experiment in Figure~\ref{fig:num_actions}. 
As explained in Section~\ref{sec:topk_offpolicy}, with $K=1$, the top-$K$ off-policy correction reduces to the standard off-policy correction. A drop of 0.66\% ViewTime was observed for $K=1$ compared with the baseline with $K=16$.  This further confirms the gain we observed shifting from the standard off-policy correction to the top-$K$ off-policy correction. Setting $K=2$ still performs worse than the production model, but the gap is reduced to 0.35\%. $K=32$ achieves similar performance as the baseline. We conducted follow up experiment which showed mildly positive gain in ViewTime (+0.15\% \emph{statistically significant}) when $K=8$.



\paragraph{Capping}
Here we consider the effect of the variance reduction techniques on the final quality of the learned recommender. 
Among the techniques discussed in Section \ref{sec:variance_reduction}, weight capping brings the biggest gain online in initial experiments.  We did not observe further metric improvements from normalized importance sampling, or TRPO \cite{schulman2015trust}. 
We conducted a regression test to study the impact of weight capping. We compare a model trained using cap $c=e^3$ (as in production model) in \eq~\eqref{eq:capping} with one trained using $c=e^5$. As we lift the restriction on the importance weight, the learned policy $\pi_\theta$ could potentially overfit to a few logged actions that accidentally receives high reward. \citet{swaminathan2015self} described a similar effect of propensity overfitting. 
During live experiment, we observe a significant drop of 0.52\% in ViewTime when the cap on importance weight was lifted. 




\section{Conclusion}
In this paper we have laid out a practical implementation of a policy gradient-based top-$K$ recommender system in use at YouTube.  We scale up REINFORCE to an action space in the orders of millions and have it stably running in a live production system. To realize the full benefits of such an approach, we have demonstrated how to address biases in logged data through incorporating a learned logging policy and a novel top-$K$ off-policy correction. We conducted extensive analysis and live experiments to measure empirically the importance of accounting for and addressing these underlying biases.  We believe these are important steps in making reinforcement learning practically impactful for recommendation and will provide a solid foundation for researchers and practitioners to explore new directions of applying RL to recommender systems. 

\section{Acknowledgements}
We thank Craig Boutilier for his valuable comments and discussions.



\bibliographystyle{ACM-Reference-Format}
\bibliography{main}


\begin{thebibliography}{50}


\ifx \showCODEN    \undefined \def \showCODEN     #1{\unskip}     \fi
\ifx \showDOI      \undefined \def \showDOI       #1{#1}\fi
\ifx \showISBNx    \undefined \def \showISBNx     #1{\unskip}     \fi
\ifx \showISBNxiii \undefined \def \showISBNxiii  #1{\unskip}     \fi
\ifx \showISSN     \undefined \def \showISSN      #1{\unskip}     \fi
\ifx \showLCCN     \undefined \def \showLCCN      #1{\unskip}     \fi
\ifx \shownote     \undefined \def \shownote      #1{#1}          \fi
\ifx \showarticletitle \undefined \def \showarticletitle #1{#1}   \fi
\ifx \showURL      \undefined \def \showURL       {\relax}        \fi
\providecommand\bibfield[2]{#2}
\providecommand\bibinfo[2]{#2}
\providecommand\natexlab[1]{#1}
\providecommand\showeprint[2][]{arXiv:#2}

\bibitem[\protect\citeauthoryear{Achiam, Held, Tamar, and Abbeel}{Achiam
  et~al\mbox{.}}{2017}]%
        {achiam2017constrained}
\bibfield{author}{\bibinfo{person}{Joshua Achiam}, \bibinfo{person}{David
  Held}, \bibinfo{person}{Aviv Tamar}, {and} \bibinfo{person}{Pieter Abbeel}.}
  \bibinfo{year}{2017}\natexlab{}.
\newblock \showarticletitle{Constrained policy optimization}.
\newblock \bibinfo{journal}{\emph{arXiv preprint arXiv:1705.10528}}
  (\bibinfo{year}{2017}).
\newblock


\bibitem[\protect\citeauthoryear{Agarwal, Hsu, Kale, Langford, Li, and
  Schapire}{Agarwal et~al\mbox{.}}{2014}]%
        {agarwal2014taming}
\bibfield{author}{\bibinfo{person}{Alekh Agarwal}, \bibinfo{person}{Daniel
  Hsu}, \bibinfo{person}{Satyen Kale}, \bibinfo{person}{John Langford},
  \bibinfo{person}{Lihong Li}, {and} \bibinfo{person}{Robert Schapire}.}
  \bibinfo{year}{2014}\natexlab{}.
\newblock \showarticletitle{Taming the monster: A fast and simple algorithm for
  contextual bandits}. In \bibinfo{booktitle}{\emph{International Conference on
  Machine Learning}}. \bibinfo{pages}{1638--1646}.
\newblock


\bibitem[\protect\citeauthoryear{Auer, Cesa-Bianchi, and Fischer}{Auer
  et~al\mbox{.}}{2002}]%
        {auer2002finite}
\bibfield{author}{\bibinfo{person}{Peter Auer}, \bibinfo{person}{Nicolo
  Cesa-Bianchi}, {and} \bibinfo{person}{Paul Fischer}.}
  \bibinfo{year}{2002}\natexlab{}.
\newblock \showarticletitle{Finite-time analysis of the multiarmed bandit
  problem}.
\newblock \bibinfo{journal}{\emph{Machine learning}} \bibinfo{volume}{47},
  \bibinfo{number}{2-3} (\bibinfo{year}{2002}), \bibinfo{pages}{235--256}.
\newblock


\bibitem[\protect\citeauthoryear{Bengio, Sen{\'e}cal, et~al\mbox{.}}{Bengio
  et~al\mbox{.}}{2003}]%
        {bengio2003quick}
\bibfield{author}{\bibinfo{person}{Yoshua Bengio},
  \bibinfo{person}{Jean-S{\'e}bastien Sen{\'e}cal}, {et~al\mbox{.}}}
  \bibinfo{year}{2003}\natexlab{}.
\newblock \showarticletitle{Quick Training of Probabilistic Neural Nets by
  Importance Sampling.}. In \bibinfo{booktitle}{\emph{AISTATS}}.
  \bibinfo{pages}{1--9}.
\newblock


\bibitem[\protect\citeauthoryear{Bennett, Lanning, et~al\mbox{.}}{Bennett
  et~al\mbox{.}}{2007}]%
        {bennett2007netflix}
\bibfield{author}{\bibinfo{person}{James Bennett}, \bibinfo{person}{Stan
  Lanning}, {et~al\mbox{.}}} \bibinfo{year}{2007}\natexlab{}.
\newblock \showarticletitle{The netflix prize}. In
  \bibinfo{booktitle}{\emph{Proceedings of KDD cup and workshop}},
  Vol.~\bibinfo{volume}{2007}. New York, NY, USA, \bibinfo{pages}{35}.
\newblock


\bibitem[\protect\citeauthoryear{Beutel, Covington, Jain, Xu, Li, Gatto, and
  Chi}{Beutel et~al\mbox{.}}{2018}]%
        {BeutelCJXLGC18}
\bibfield{author}{\bibinfo{person}{Alex Beutel}, \bibinfo{person}{Paul
  Covington}, \bibinfo{person}{Sagar Jain}, \bibinfo{person}{Can Xu},
  \bibinfo{person}{Jia Li}, \bibinfo{person}{Vince Gatto}, {and}
  \bibinfo{person}{Ed~H Chi}.} \bibinfo{year}{2018}\natexlab{}.
\newblock \showarticletitle{Latent Cross: Making Use of Context in Recurrent
  Recommender Systems}. In \bibinfo{booktitle}{\emph{WSDM}}. ACM,
  \bibinfo{pages}{46--54}.
\newblock


\bibitem[\protect\citeauthoryear{Bottou}{Bottou}{2010}]%
        {bottou2010large}
\bibfield{author}{\bibinfo{person}{L{\'e}on Bottou}.}
  \bibinfo{year}{2010}\natexlab{}.
\newblock \showarticletitle{Large-scale machine learning with stochastic
  gradient descent}.
\newblock In \bibinfo{booktitle}{\emph{Proceedings of COMPSTAT'2010}}.
  \bibinfo{publisher}{Springer}, \bibinfo{pages}{177--186}.
\newblock


\bibitem[\protect\citeauthoryear{Bottou, Peters, Qui{\~n}onero-Candela,
  Charles, Chickering, Portugaly, Ray, Simard, and Snelson}{Bottou
  et~al\mbox{.}}{2013}]%
        {bottou2013counterfactual}
\bibfield{author}{\bibinfo{person}{L{\'e}on Bottou}, \bibinfo{person}{Jonas
  Peters}, \bibinfo{person}{Joaquin Qui{\~n}onero-Candela},
  \bibinfo{person}{Denis~X Charles}, \bibinfo{person}{D~Max Chickering},
  \bibinfo{person}{Elon Portugaly}, \bibinfo{person}{Dipankar Ray},
  \bibinfo{person}{Patrice Simard}, {and} \bibinfo{person}{Ed Snelson}.}
  \bibinfo{year}{2013}\natexlab{}.
\newblock \showarticletitle{Counterfactual reasoning and learning systems: The
  example of computational advertising}.
\newblock \bibinfo{journal}{\emph{The Journal of Machine Learning Research}}
  \bibinfo{volume}{14}, \bibinfo{number}{1} (\bibinfo{year}{2013}),
  \bibinfo{pages}{3207--3260}.
\newblock


\bibitem[\protect\citeauthoryear{Chapelle and Li}{Chapelle and Li}{2011}]%
        {chapelle2011empirical}
\bibfield{author}{\bibinfo{person}{Olivier Chapelle} {and}
  \bibinfo{person}{Lihong Li}.} \bibinfo{year}{2011}\natexlab{}.
\newblock \showarticletitle{An empirical evaluation of thompson sampling}. In
  \bibinfo{booktitle}{\emph{Advances in neural information processing
  systems}}. \bibinfo{pages}{2249--2257}.
\newblock


\bibitem[\protect\citeauthoryear{Chung, Gulcehre, Cho, and Bengio}{Chung
  et~al\mbox{.}}{2014}]%
        {chung2014empirical}
\bibfield{author}{\bibinfo{person}{Junyoung Chung}, \bibinfo{person}{Caglar
  Gulcehre}, \bibinfo{person}{KyungHyun Cho}, {and} \bibinfo{person}{Yoshua
  Bengio}.} \bibinfo{year}{2014}\natexlab{}.
\newblock \showarticletitle{Empirical evaluation of gated recurrent neural
  networks on sequence modeling}.
\newblock \bibinfo{journal}{\emph{arXiv preprint arXiv:1412.3555}}
  (\bibinfo{year}{2014}).
\newblock


\bibitem[\protect\citeauthoryear{Covington, Adams, and Sargin}{Covington
  et~al\mbox{.}}{2016}]%
        {covington2016deep}
\bibfield{author}{\bibinfo{person}{Paul Covington}, \bibinfo{person}{Jay
  Adams}, {and} \bibinfo{person}{Emre Sargin}.}
  \bibinfo{year}{2016}\natexlab{}.
\newblock \showarticletitle{Deep neural networks for youtube recommendations}.
  In \bibinfo{booktitle}{\emph{Proceedings of the 10th ACM Conference on
  Recommender Systems}}. ACM, \bibinfo{pages}{191--198}.
\newblock


\bibitem[\protect\citeauthoryear{Daw, O'doherty, Dayan, Seymour, and Dolan}{Daw
  et~al\mbox{.}}{2006}]%
        {daw2006cortical}
\bibfield{author}{\bibinfo{person}{Nathaniel~D Daw}, \bibinfo{person}{John~P
  O'doherty}, \bibinfo{person}{Peter Dayan}, \bibinfo{person}{Ben Seymour},
  {and} \bibinfo{person}{Raymond~J Dolan}.} \bibinfo{year}{2006}\natexlab{}.
\newblock \showarticletitle{Cortical substrates for exploratory decisions in
  humans}.
\newblock \bibinfo{journal}{\emph{Nature}} \bibinfo{volume}{441},
  \bibinfo{number}{7095} (\bibinfo{year}{2006}), \bibinfo{pages}{876}.
\newblock


\bibitem[\protect\citeauthoryear{Gilotte, Calauz{\`e}nes, Nedelec, Abraham, and
  Doll{\'e}}{Gilotte et~al\mbox{.}}{2018}]%
        {gilotte2018offline}
\bibfield{author}{\bibinfo{person}{Alexandre Gilotte},
  \bibinfo{person}{Cl{\'e}ment Calauz{\`e}nes}, \bibinfo{person}{Thomas
  Nedelec}, \bibinfo{person}{Alexandre Abraham}, {and} \bibinfo{person}{Simon
  Doll{\'e}}.} \bibinfo{year}{2018}\natexlab{}.
\newblock \showarticletitle{Offline A/B testing for Recommender Systems}. In
  \bibinfo{booktitle}{\emph{WSDM}}. ACM, \bibinfo{pages}{198--206}.
\newblock


\bibitem[\protect\citeauthoryear{Guo, Kumar, Choromanski, and Simcha}{Guo
  et~al\mbox{.}}{2016}]%
        {guo2016quantization}
\bibfield{author}{\bibinfo{person}{Ruiqi Guo}, \bibinfo{person}{Sanjiv Kumar},
  \bibinfo{person}{Krzysztof Choromanski}, {and} \bibinfo{person}{David
  Simcha}.} \bibinfo{year}{2016}\natexlab{}.
\newblock \showarticletitle{Quantization based fast inner product search}. In
  \bibinfo{booktitle}{\emph{Artificial Intelligence and Statistics}}.
  \bibinfo{pages}{482--490}.
\newblock


\bibitem[\protect\citeauthoryear{Hansen and Ostermeier}{Hansen and
  Ostermeier}{2001}]%
        {hansen2001completely}
\bibfield{author}{\bibinfo{person}{Nikolaus Hansen} {and}
  \bibinfo{person}{Andreas Ostermeier}.} \bibinfo{year}{2001}\natexlab{}.
\newblock \showarticletitle{Completely derandomized self-adaptation in
  evolution strategies}.
\newblock \bibinfo{journal}{\emph{Evolutionary computation}}
  \bibinfo{volume}{9}, \bibinfo{number}{2} (\bibinfo{year}{2001}),
  \bibinfo{pages}{159--195}.
\newblock


\bibitem[\protect\citeauthoryear{He, Liao, Zhang, Nie, Hu, and Chua}{He
  et~al\mbox{.}}{2017}]%
        {he2017neural}
\bibfield{author}{\bibinfo{person}{Xiangnan He}, \bibinfo{person}{Lizi Liao},
  \bibinfo{person}{Hanwang Zhang}, \bibinfo{person}{Liqiang Nie},
  \bibinfo{person}{Xia Hu}, {and} \bibinfo{person}{Tat-Seng Chua}.}
  \bibinfo{year}{2017}\natexlab{}.
\newblock \showarticletitle{Neural collaborative filtering}. In
  \bibinfo{booktitle}{\emph{Proceedings of the 26th International Conference on
  World Wide Web}}. International World Wide Web Conferences Steering
  Committee, \bibinfo{pages}{173--182}.
\newblock


\bibitem[\protect\citeauthoryear{Hidasi, Karatzoglou, Baltrunas, and
  Tikk}{Hidasi et~al\mbox{.}}{2015}]%
        {hidasi2015session}
\bibfield{author}{\bibinfo{person}{Bal{\'a}zs Hidasi},
  \bibinfo{person}{Alexandros Karatzoglou}, \bibinfo{person}{Linas Baltrunas},
  {and} \bibinfo{person}{Domonkos Tikk}.} \bibinfo{year}{2015}\natexlab{}.
\newblock \showarticletitle{Session-based recommendations with recurrent neural
  networks}.
\newblock \bibinfo{journal}{\emph{arXiv preprint arXiv:1511.06939}}
  (\bibinfo{year}{2015}).
\newblock


\bibitem[\protect\citeauthoryear{Hochreiter and Schmidhuber}{Hochreiter and
  Schmidhuber}{1997}]%
        {hochreiter1997long}
\bibfield{author}{\bibinfo{person}{Sepp Hochreiter} {and}
  \bibinfo{person}{J{\"u}rgen Schmidhuber}.} \bibinfo{year}{1997}\natexlab{}.
\newblock \showarticletitle{Long short-term memory}.
\newblock \bibinfo{journal}{\emph{Neural computation}} \bibinfo{volume}{9},
  \bibinfo{number}{8} (\bibinfo{year}{1997}), \bibinfo{pages}{1735--1780}.
\newblock


\bibitem[\protect\citeauthoryear{Hu, Koren, and Volinsky}{Hu
  et~al\mbox{.}}{2008}]%
        {hu2008collaborative}
\bibfield{author}{\bibinfo{person}{Yifan Hu}, \bibinfo{person}{Yehuda Koren},
  {and} \bibinfo{person}{Chris Volinsky}.} \bibinfo{year}{2008}\natexlab{}.
\newblock \showarticletitle{Collaborative filtering for implicit feedback
  datasets}. In \bibinfo{booktitle}{\emph{ICDM}}. Ieee,
  \bibinfo{pages}{263--272}.
\newblock


\bibitem[\protect\citeauthoryear{Jing and Smola}{Jing and Smola}{2017}]%
        {jing2017neural}
\bibfield{author}{\bibinfo{person}{How Jing} {and} \bibinfo{person}{Alexander~J
  Smola}.} \bibinfo{year}{2017}\natexlab{}.
\newblock \showarticletitle{Neural survival recommender}. In
  \bibinfo{booktitle}{\emph{WSDM}}. ACM, \bibinfo{pages}{515--524}.
\newblock


\bibitem[\protect\citeauthoryear{Joachims, Swaminathan, and Schnabel}{Joachims
  et~al\mbox{.}}{2017}]%
        {joachims2017unbiased}
\bibfield{author}{\bibinfo{person}{Thorsten Joachims}, \bibinfo{person}{Adith
  Swaminathan}, {and} \bibinfo{person}{Tobias Schnabel}.}
  \bibinfo{year}{2017}\natexlab{}.
\newblock \showarticletitle{Unbiased learning-to-rank with biased feedback}. In
  \bibinfo{booktitle}{\emph{Proceedings of the Tenth ACM International
  Conference on Web Search and Data Mining}}. ACM, \bibinfo{pages}{781--789}.
\newblock


\bibitem[\protect\citeauthoryear{Kober, Bagnell, and Peters}{Kober
  et~al\mbox{.}}{2013}]%
        {kober2013reinforcement}
\bibfield{author}{\bibinfo{person}{Jens Kober}, \bibinfo{person}{J~Andrew
  Bagnell}, {and} \bibinfo{person}{Jan Peters}.}
  \bibinfo{year}{2013}\natexlab{}.
\newblock \showarticletitle{Reinforcement learning in robotics: A survey}.
\newblock \bibinfo{journal}{\emph{The International Journal of Robotics
  Research}} \bibinfo{volume}{32}, \bibinfo{number}{11} (\bibinfo{year}{2013}),
  \bibinfo{pages}{1238--1274}.
\newblock


\bibitem[\protect\citeauthoryear{Lapin, Hein, and Schiele}{Lapin
  et~al\mbox{.}}{2016}]%
        {lapin2016loss}
\bibfield{author}{\bibinfo{person}{Maksim Lapin}, \bibinfo{person}{Matthias
  Hein}, {and} \bibinfo{person}{Bernt Schiele}.}
  \bibinfo{year}{2016}\natexlab{}.
\newblock \showarticletitle{Loss functions for top-k error: Analysis and
  insights}. In \bibinfo{booktitle}{\emph{Proceedings of the IEEE Conference on
  Computer Vision and Pattern Recognition}}. \bibinfo{pages}{1468--1477}.
\newblock


\bibitem[\protect\citeauthoryear{Laurent and von Brecht}{Laurent and von
  Brecht}{2016}]%
        {laurent2016recurrent}
\bibfield{author}{\bibinfo{person}{Thomas Laurent} {and} \bibinfo{person}{James
  von Brecht}.} \bibinfo{year}{2016}\natexlab{}.
\newblock \showarticletitle{A recurrent neural network without chaos}.
\newblock \bibinfo{journal}{\emph{arXiv preprint arXiv:1612.06212}}
  (\bibinfo{year}{2016}).
\newblock


\bibitem[\protect\citeauthoryear{Levine, Finn, Darrell, and Abbeel}{Levine
  et~al\mbox{.}}{2016}]%
        {levine2016end}
\bibfield{author}{\bibinfo{person}{Sergey Levine}, \bibinfo{person}{Chelsea
  Finn}, \bibinfo{person}{Trevor Darrell}, {and} \bibinfo{person}{Pieter
  Abbeel}.} \bibinfo{year}{2016}\natexlab{}.
\newblock \showarticletitle{End-to-end training of deep visuomotor policies}.
\newblock \bibinfo{journal}{\emph{JMLR}} \bibinfo{volume}{17},
  \bibinfo{number}{1} (\bibinfo{year}{2016}), \bibinfo{pages}{1334--1373}.
\newblock


\bibitem[\protect\citeauthoryear{Levine and Koltun}{Levine and Koltun}{2013}]%
        {levine2013guided}
\bibfield{author}{\bibinfo{person}{Sergey Levine} {and}
  \bibinfo{person}{Vladlen Koltun}.} \bibinfo{year}{2013}\natexlab{}.
\newblock \showarticletitle{Guided policy search}. In
  \bibinfo{booktitle}{\emph{International Conference on Machine Learning}}.
  \bibinfo{pages}{1--9}.
\newblock


\bibitem[\protect\citeauthoryear{Li, Chu, Langford, and Schapire}{Li
  et~al\mbox{.}}{2010}]%
        {li2010contextual}
\bibfield{author}{\bibinfo{person}{Lihong Li}, \bibinfo{person}{Wei Chu},
  \bibinfo{person}{John Langford}, {and} \bibinfo{person}{Robert~E Schapire}.}
  \bibinfo{year}{2010}\natexlab{}.
\newblock \showarticletitle{A contextual-bandit approach to personalized news
  article recommendation}. In \bibinfo{booktitle}{\emph{Proceedings of the 19th
  international conference on World wide web}}. ACM, \bibinfo{pages}{661--670}.
\newblock


\bibitem[\protect\citeauthoryear{Mary, Gaudel, and Preux}{Mary
  et~al\mbox{.}}{2015}]%
        {mary2015bandits}
\bibfield{author}{\bibinfo{person}{J{\'e}r{\'e}mie Mary},
  \bibinfo{person}{Romaric Gaudel}, {and} \bibinfo{person}{Philippe Preux}.}
  \bibinfo{year}{2015}\natexlab{}.
\newblock \showarticletitle{Bandits and recommender systems}. In
  \bibinfo{booktitle}{\emph{International Workshop on Machine Learning,
  Optimization and Big Data}}. Springer, \bibinfo{pages}{325--336}.
\newblock


\bibitem[\protect\citeauthoryear{Mnih, Badia, Mirza, Graves, Lillicrap, Harley,
  Silver, and Kavukcuoglu}{Mnih et~al\mbox{.}}{2016}]%
        {mnih2016asynchronous}
\bibfield{author}{\bibinfo{person}{Volodymyr Mnih},
  \bibinfo{person}{Adria~Puigdomenech Badia}, \bibinfo{person}{Mehdi Mirza},
  \bibinfo{person}{Alex Graves}, \bibinfo{person}{Timothy Lillicrap},
  \bibinfo{person}{Tim Harley}, \bibinfo{person}{David Silver}, {and}
  \bibinfo{person}{Koray Kavukcuoglu}.} \bibinfo{year}{2016}\natexlab{}.
\newblock \showarticletitle{Asynchronous methods for deep reinforcement
  learning}. In \bibinfo{booktitle}{\emph{International conference on machine
  learning}}. \bibinfo{pages}{1928--1937}.
\newblock


\bibitem[\protect\citeauthoryear{Mnih, Kavukcuoglu, Silver, Graves, Antonoglou,
  Wierstra, and Riedmiller}{Mnih et~al\mbox{.}}{2013}]%
        {mnih2013playing}
\bibfield{author}{\bibinfo{person}{Volodymyr Mnih}, \bibinfo{person}{Koray
  Kavukcuoglu}, \bibinfo{person}{David Silver}, \bibinfo{person}{Alex Graves},
  \bibinfo{person}{Ioannis Antonoglou}, \bibinfo{person}{Daan Wierstra}, {and}
  \bibinfo{person}{Martin Riedmiller}.} \bibinfo{year}{2013}\natexlab{}.
\newblock \showarticletitle{Playing atari with deep reinforcement learning}.
\newblock \bibinfo{journal}{\emph{arXiv preprint arXiv:1312.5602}}
  (\bibinfo{year}{2013}).
\newblock


\bibitem[\protect\citeauthoryear{Munos, Stepleton, Harutyunyan, and
  Bellemare}{Munos et~al\mbox{.}}{2016}]%
        {munos2016safe}
\bibfield{author}{\bibinfo{person}{R{\'e}mi Munos}, \bibinfo{person}{Tom
  Stepleton}, \bibinfo{person}{Anna Harutyunyan}, {and} \bibinfo{person}{Marc
  Bellemare}.} \bibinfo{year}{2016}\natexlab{}.
\newblock \showarticletitle{Safe and efficient off-policy reinforcement
  learning}. In \bibinfo{booktitle}{\emph{Advances in Neural Information
  Processing Systems}}. \bibinfo{pages}{1054--1062}.
\newblock


\bibitem[\protect\citeauthoryear{Owen}{Owen}{2013}]%
        {mcbook}
\bibfield{author}{\bibinfo{person}{Art~B. Owen}.}
  \bibinfo{year}{2013}\natexlab{}.
\newblock \bibinfo{booktitle}{\emph{Monte Carlo theory, methods and examples}}.
\newblock


\bibitem[\protect\citeauthoryear{Precup}{Precup}{2000}]%
        {precup2000eligibility}
\bibfield{author}{\bibinfo{person}{Doina Precup}.}
  \bibinfo{year}{2000}\natexlab{}.
\newblock \showarticletitle{Eligibility traces for off-policy policy
  evaluation}.
\newblock \bibinfo{journal}{\emph{Computer Science Department Faculty
  Publication Series}} (\bibinfo{year}{2000}), \bibinfo{pages}{80}.
\newblock


\bibitem[\protect\citeauthoryear{Precup, Sutton, and Dasgupta}{Precup
  et~al\mbox{.}}{2001}]%
        {precup2001off}
\bibfield{author}{\bibinfo{person}{Doina Precup}, \bibinfo{person}{Richard~S
  Sutton}, {and} \bibinfo{person}{Sanjoy Dasgupta}.}
  \bibinfo{year}{2001}\natexlab{}.
\newblock \showarticletitle{Off-policy temporal-difference learning with
  function approximation}. In \bibinfo{booktitle}{\emph{ICML}}.
  \bibinfo{pages}{417--424}.
\newblock


\bibitem[\protect\citeauthoryear{Schnabel, Bennett, Dumais, and
  Joachims}{Schnabel et~al\mbox{.}}{2018}]%
        {schnabel2018short}
\bibfield{author}{\bibinfo{person}{Tobias Schnabel}, \bibinfo{person}{Paul~N
  Bennett}, \bibinfo{person}{Susan~T Dumais}, {and} \bibinfo{person}{Thorsten
  Joachims}.} \bibinfo{year}{2018}\natexlab{}.
\newblock \showarticletitle{Short-term satisfaction and long-term coverage:
  Understanding how users tolerate algorithmic exploration}. In
  \bibinfo{booktitle}{\emph{Proceedings of the Eleventh ACM International
  Conference on Web Search and Data Mining}}. ACM, \bibinfo{pages}{513--521}.
\newblock


\bibitem[\protect\citeauthoryear{Schulman, Levine, Abbeel, Jordan, and
  Moritz}{Schulman et~al\mbox{.}}{2015}]%
        {schulman2015trust}
\bibfield{author}{\bibinfo{person}{John Schulman}, \bibinfo{person}{Sergey
  Levine}, \bibinfo{person}{Pieter Abbeel}, \bibinfo{person}{Michael Jordan},
  {and} \bibinfo{person}{Philipp Moritz}.} \bibinfo{year}{2015}\natexlab{}.
\newblock \showarticletitle{Trust region policy optimization}. In
  \bibinfo{booktitle}{\emph{International Conference on Machine Learning}}.
  \bibinfo{pages}{1889--1897}.
\newblock


\bibitem[\protect\citeauthoryear{Sedhain, Menon, Sanner, and Xie}{Sedhain
  et~al\mbox{.}}{2015}]%
        {sedhain2015autorec}
\bibfield{author}{\bibinfo{person}{Suvash Sedhain},
  \bibinfo{person}{Aditya~Krishna Menon}, \bibinfo{person}{Scott Sanner}, {and}
  \bibinfo{person}{Lexing Xie}.} \bibinfo{year}{2015}\natexlab{}.
\newblock \showarticletitle{Autorec: Autoencoders meet collaborative
  filtering}. In \bibinfo{booktitle}{\emph{Proceedings of the 24th
  International Conference on World Wide Web}}. ACM, \bibinfo{pages}{111--112}.
\newblock


\bibitem[\protect\citeauthoryear{Silver, Huang, Maddison, Guez, Sifre, Van
  Den~Driessche, Schrittwieser, Antonoglou, Panneershelvam, Lanctot,
  et~al\mbox{.}}{Silver et~al\mbox{.}}{2016}]%
        {silver2016mastering}
\bibfield{author}{\bibinfo{person}{David Silver}, \bibinfo{person}{Aja Huang},
  \bibinfo{person}{Chris~J Maddison}, \bibinfo{person}{Arthur Guez},
  \bibinfo{person}{Laurent Sifre}, \bibinfo{person}{George Van Den~Driessche},
  \bibinfo{person}{Julian Schrittwieser}, \bibinfo{person}{Ioannis Antonoglou},
  \bibinfo{person}{Veda Panneershelvam}, \bibinfo{person}{Marc Lanctot},
  {et~al\mbox{.}}} \bibinfo{year}{2016}\natexlab{}.
\newblock \showarticletitle{Mastering the game of Go with deep neural networks
  and tree search}.
\newblock \bibinfo{journal}{\emph{nature}} \bibinfo{volume}{529},
  \bibinfo{number}{7587} (\bibinfo{year}{2016}), \bibinfo{pages}{484}.
\newblock


\bibitem[\protect\citeauthoryear{Strehl, Langford, Li, and Kakade}{Strehl
  et~al\mbox{.}}{2010}]%
        {strehl2010learning}
\bibfield{author}{\bibinfo{person}{Alex Strehl}, \bibinfo{person}{John
  Langford}, \bibinfo{person}{Lihong Li}, {and} \bibinfo{person}{Sham~M
  Kakade}.} \bibinfo{year}{2010}\natexlab{}.
\newblock \showarticletitle{Learning from logged implicit exploration data}. In
  \bibinfo{booktitle}{\emph{Advances in Neural Information Processing
  Systems}}. \bibinfo{pages}{2217--2225}.
\newblock


\bibitem[\protect\citeauthoryear{Sutton, Barto, et~al\mbox{.}}{Sutton
  et~al\mbox{.}}{1998}]%
        {sutton1998reinforcement}
\bibfield{author}{\bibinfo{person}{Richard~S Sutton}, \bibinfo{person}{Andrew~G
  Barto}, {et~al\mbox{.}}} \bibinfo{year}{1998}\natexlab{}.
\newblock \bibinfo{booktitle}{\emph{Reinforcement learning: An introduction}}.
\newblock \bibinfo{publisher}{MIT press}.
\newblock


\bibitem[\protect\citeauthoryear{Sutton, McAllester, Singh, and Mansour}{Sutton
  et~al\mbox{.}}{2000}]%
        {sutton2000policy}
\bibfield{author}{\bibinfo{person}{Richard~S Sutton}, \bibinfo{person}{David~A
  McAllester}, \bibinfo{person}{Satinder~P Singh}, {and}
  \bibinfo{person}{Yishay Mansour}.} \bibinfo{year}{2000}\natexlab{}.
\newblock \showarticletitle{Policy gradient methods for reinforcement learning
  with function approximation}. In \bibinfo{booktitle}{\emph{Advances in neural
  information processing systems}}. \bibinfo{pages}{1057--1063}.
\newblock


\bibitem[\protect\citeauthoryear{Swaminathan and Joachims}{Swaminathan and
  Joachims}{2015a}]%
        {swaminathan2015batch}
\bibfield{author}{\bibinfo{person}{Adith Swaminathan} {and}
  \bibinfo{person}{Thorsten Joachims}.} \bibinfo{year}{2015}\natexlab{a}.
\newblock \showarticletitle{Batch learning from logged bandit feedback through
  counterfactual risk minimization.}
\newblock \bibinfo{journal}{\emph{Journal of Machine Learning Research}}
  \bibinfo{volume}{16}, \bibinfo{number}{1} (\bibinfo{year}{2015}),
  \bibinfo{pages}{1731--1755}.
\newblock


\bibitem[\protect\citeauthoryear{Swaminathan and Joachims}{Swaminathan and
  Joachims}{2015b}]%
        {swaminathan2015self}
\bibfield{author}{\bibinfo{person}{Adith Swaminathan} {and}
  \bibinfo{person}{Thorsten Joachims}.} \bibinfo{year}{2015}\natexlab{b}.
\newblock \showarticletitle{The self-normalized estimator for counterfactual
  learning}. In \bibinfo{booktitle}{\emph{Advances in Neural Information
  Processing Systems}}. \bibinfo{pages}{3231--3239}.
\newblock


\bibitem[\protect\citeauthoryear{Swaminathan, Krishnamurthy, Agarwal, Dudik,
  Langford, Jose, and Zitouni}{Swaminathan et~al\mbox{.}}{2017}]%
        {swaminathan2017off}
\bibfield{author}{\bibinfo{person}{Adith Swaminathan}, \bibinfo{person}{Akshay
  Krishnamurthy}, \bibinfo{person}{Alekh Agarwal}, \bibinfo{person}{Miro
  Dudik}, \bibinfo{person}{John Langford}, \bibinfo{person}{Damien Jose}, {and}
  \bibinfo{person}{Imed Zitouni}.} \bibinfo{year}{2017}\natexlab{}.
\newblock \showarticletitle{Off-policy evaluation for slate recommendation}. In
  \bibinfo{booktitle}{\emph{Advances in Neural Information Processing
  Systems}}.
\newblock


\bibitem[\protect\citeauthoryear{Tan, Xu, and Liu}{Tan et~al\mbox{.}}{2016}]%
        {tan2016improved}
\bibfield{author}{\bibinfo{person}{Yong~Kiam Tan}, \bibinfo{person}{Xinxing
  Xu}, {and} \bibinfo{person}{Yong Liu}.} \bibinfo{year}{2016}\natexlab{}.
\newblock \showarticletitle{Improved recurrent neural networks for
  session-based recommendations}. In \bibinfo{booktitle}{\emph{Proceedings of
  the 1st Workshop on Deep Learning for Recommender Systems}}. ACM,
  \bibinfo{pages}{17--22}.
\newblock


\bibitem[\protect\citeauthoryear{Tesauro}{Tesauro}{1995}]%
        {tesauro1995temporal}
\bibfield{author}{\bibinfo{person}{Gerald Tesauro}.}
  \bibinfo{year}{1995}\natexlab{}.
\newblock \showarticletitle{Temporal difference learning and TD-Gammon}.
\newblock \bibinfo{journal}{\emph{Commun. ACM}} \bibinfo{volume}{38},
  \bibinfo{number}{3} (\bibinfo{year}{1995}), \bibinfo{pages}{58--68}.
\newblock


\bibitem[\protect\citeauthoryear{Thomas and Brunskill}{Thomas and
  Brunskill}{2016}]%
        {thomas2016data}
\bibfield{author}{\bibinfo{person}{Philip Thomas} {and} \bibinfo{person}{Emma
  Brunskill}.} \bibinfo{year}{2016}\natexlab{}.
\newblock \showarticletitle{Data-efficient off-policy policy evaluation for
  reinforcement learning}. In \bibinfo{booktitle}{\emph{ICML}}.
  \bibinfo{pages}{2139--2148}.
\newblock


\bibitem[\protect\citeauthoryear{Williams}{Williams}{1992}]%
        {williams1992simple}
\bibfield{author}{\bibinfo{person}{Ronald~J Williams}.}
  \bibinfo{year}{1992}\natexlab{}.
\newblock \showarticletitle{Simple statistical gradient-following algorithms
  for connectionist reinforcement learning}.
\newblock \bibinfo{journal}{\emph{Machine learning}} \bibinfo{volume}{8},
  \bibinfo{number}{3-4} (\bibinfo{year}{1992}), \bibinfo{pages}{229--256}.
\newblock


\bibitem[\protect\citeauthoryear{Wu, Ahmed, Beutel, Smola, and Jing}{Wu
  et~al\mbox{.}}{2017}]%
        {WuABSJ17}
\bibfield{author}{\bibinfo{person}{Chao-Yuan Wu}, \bibinfo{person}{Amr Ahmed},
  \bibinfo{person}{Alex Beutel}, \bibinfo{person}{Alexander~J Smola}, {and}
  \bibinfo{person}{How Jing}.} \bibinfo{year}{2017}\natexlab{}.
\newblock \showarticletitle{Recurrent recommender networks}. In
  \bibinfo{booktitle}{\emph{Proceedings of the tenth ACM international
  conference on web search and data mining}}. ACM, \bibinfo{pages}{495--503}.
\newblock


\bibitem[\protect\citeauthoryear{Zhao, Xia, Zhang, Ding, Yin, and Tang}{Zhao
  et~al\mbox{.}}{2018}]%
        {zhao2018deep}
\bibfield{author}{\bibinfo{person}{Xiangyu Zhao}, \bibinfo{person}{Long Xia},
  \bibinfo{person}{Liang Zhang}, \bibinfo{person}{Zhuoye Ding},
  \bibinfo{person}{Dawei Yin}, {and} \bibinfo{person}{Jiliang Tang}.}
  \bibinfo{year}{2018}\natexlab{}.
\newblock \showarticletitle{Deep Reinforcement Learning for Page-wise
  Recommendations}.
\newblock \bibinfo{journal}{\emph{arXiv preprint arXiv:1805.02343}}
  (\bibinfo{year}{2018}).
\newblock


\end{thebibliography}

\end{document}